\pdfoutput=1
\documentclass[11pt]{article}

\usepackage[table,xcdraw]{xcolor}

\usepackage[]{emnlp2021}

\usepackage{times}
\usepackage{latexsym}

\usepackage[T1]{fontenc}
\usepackage[utf8]{inputenc}

\usepackage{microtype}

\usepackage{float}
\usepackage{subfig}
\usepackage{tabularx}
\usepackage{hyperref}
\usepackage{graphicx}
\usepackage{url}
\usepackage{fancyhdr}
\usepackage{booktabs}

\title{Tracking Turbulence Through Financial News During COVID-19}

\author{Philip Hossu \and Natalie Parde \\
  University of Illinois at Chicago \\
  Department of Computer Science \\
  \texttt{\{phossu2, parde\}@uic.edu} }

\begin{document}
\maketitle
\begin{abstract}

Grave human toll notwithstanding, the COVID-19 pandemic created uniquely unstable conditions in financial markets. In this work we uncover and discuss relationships involving sentiment in financial publications during the 2020 pandemic-motivated U.S. financial crash. First, we introduce a set of expert annotations of financial sentiment for articles from major American financial news publishers. After an exploratory data analysis, we then describe a CNN-based architecture to address the task of predicting financial sentiment in this anomalous, tumultuous setting.  Our best performing model achieves a maximum weighted F1 score of 0.746, establishing a strong performance benchmark. Using predictions from our top performing model, we close by conducting a statistical correlation study with real stock market data, finding interesting and strong relationships between financial news and the S\&P 500 index, trading volume, market volatility, and different single-factor ETFs. 
\end{abstract}

\section{Introduction}
Recent advancements in fundamental natural language understanding problems \cite{zhang-etal-2019-ernie,2020t5,Clark2020ELECTRA:} have generated renewed interest in their downstream deployment in many applications; one such example is extrapolating meaning from large quantities of financial publications.  The year 2020 and its accompanying financial unpredictability presents an interesting case study and test bed for financial sentiment models that rely on complex inferences.
In this work we tackle that problem, and uncover numerous interesting and important relationships involving sentiment in financial publications during the 2020 pandemic-related U.S. financial crash.  This work is unique as we focus not on lexicon scoring, nor developing a trading strategy during normal market conditions, but instead on analyzing the relationship between sentiment and market movement during a financial crisis --- a subject seldom studied due to the infrequency of financial crashes and pandemics. Our key contributions are as follows:

\begin{itemize}
    \item We collect a set of expert annotations for 800 financial article titles and descriptions with COVID-19-specific vocabulary, labeled for negative, neutral, or positive sentiment.
    \item We introduce CNN-based models to distinguish between sentiment categories in this challenging dataset, achieving a maximum weighted F1 score of 0.746.
    \item We perform correlation studies between our top model's predictions and popular market measures: S\&P 500 Average, S\&P 500 Trading Volume, Cboe Volatility Index, Single-Factor ETFs, and FAANG+ stocks.  
\end{itemize}

Our experiments yield intriguing findings, including a strong relationship between predicted market sentiment and the S\&P 500 Average and a notably lower correlation with value stocks compared to other factors.

\section{Related Work}

Sentiment analysis is a popular NLP task \cite{ribeiro_sentibench_2016} that has been employed in a variety of practical fields, ranging from product review analyses for marketing purposes \cite{hu_mining_2004}, to tracking online voter sentiment \cite{tumasjan_election_2011}, as well as to financial markets. Current methods for extracting sentiment from bodies of text fall primarily into two categories: lexicon-based and machine learning methods. Lexicon-based approaches ultimately produce dictionaries containing key/value pairs for scoring text. Although simplistic, they have the added benefit of being extremely efficient, requiring user programs to do little more than load a dictionary into memory and perform linear searches. 

However, lexicon-based methods tend to fall short compared to modern deep learning techniques, especially in domains with specialized vocabularies. Machine learning methods often outperform lexicons in financial text applications \cite{man_financial_2019}. Financial texts not only contain highly specialized language referring to specific actions and outcomes (eg., \textit{[Company] prices senior notes} announces a type of bond offering), but also exhibit language that can have very different meaning compared to general settings (eg., ``bull'' may typically have a neutral or negative connotation, but is overwhelmingly positive in a financial context). 

A recent surge of work in financial sentiment analysis emerged during SEMEVAL 2017's Task 5 \cite{semeval_2017}, in which participants were asked to create systems capable of predicting article sentiment on a continuous scale from -1 (bearish/negative/price-down) to 1 (bullish/positive/price-up), given a dataset of 1647 annotated financial statements and headlines. The top two performing systems \cite{mansar_fortia-fbk_2017,kar_ritual-uh_2017} performed within 0.001 points of each other, yet used vastly different techniques. \citet{mansar_fortia-fbk_2017} utilized three primary input features: pre-trained GloVe embeddings \cite{pennington_glove}, DepecheMood scores \cite{staiano_depechemood}, and VADER scores \cite{vader_paper}.  They paired these with a CNN-based architecture, leveraging the following layers in order: convolution, max pooling, dropout layer, fully connected layer, dropout layer, and fully connected layer. 

Conversely, \citet{kar_ritual-uh_2017} used pre-trained Word2Vec embeddings \cite{mikolov_efficient_2013} as input separately to a CNN and a bidirectional gated recurrent unit (BiGRU), and also generated a set of hand-crafted features which were fed to multiple fully-connected layers. All three of these model outputs were concatenated, before being input to another fully-connected layer which ultimately issued a prediction. We take inspiration from the former paper \cite{mansar_fortia-fbk_2017}, and experiment with a similar CNN-based approach. 

\section{Data Sources}

We collected financial articles using the NewsFilter.IO API.\footnote{\url{http://newsfilter.io}} This service provides straightforward and high quality data access for online articles from most major American publishers of financial news. Articles are provided in a JSON dictionary format with fields corresponding to title, description, publisher name, published time, and stock tickers mentioned, among others. We refer to this dataset from here onward as \textsc{Fin-Cov-News}.

To collect market and share prices, we downloaded publicly available data from Yahoo Finance.\footnote{\url{https://finance.yahoo.com/}} Yahoo Finance allows users to download daily price data given inputs of a specific stock or instrument symbol and date range.  

\subsection{Basic Statistics}

We collected data from January 1 to December 31, 2020. This yielded 192,154 articles from the following sources: Wall Street Journal, Bloomberg, SeekingAlpha, PR Newswire, CNBC, and Reuters. Considering only article titles, the dataset contains 2,069,244 words, or 1,660,842 words without common stopwords \cite{nltk_book}, with a vocabulary size of 118,719 unique words. 

\begin{table}[t]
\centering
\small
\begin{tabular}{@{}lrlr@{}}
\toprule
Word        & Occurrences & Word     & Occurrences \\ \midrule
coronavirus & 12427       & results  & 5101        \\
u.s.        & 12198       & reports  & 4966        \\ 
new         & 11620       & billion  & 4913        \\
says        & 10839       & china    & 4872        \\
market      & 7808        & stocks   & 4835        \\
announces   & 7468        & dividend & 4715        \\
2020        & 7282        & virus    & 4232        \\
covid-19    & 6914        & oil      & 4041        \\
global      & 6659        & declares & 3948        \\ \bottomrule
\end{tabular}
\caption{Most common words appearing in titles in \textsc{Fin-Cov-News}.}
\label{s2:tab1}
\end{table}

The twenty most common words in titles (minus punctuation) are shown in Table \ref{s2:tab1}. The (mean, min, max) set for titles is (8.64, 1, 54), and for article descriptions is (27.35, 1, 1309). As expected, a significant number of COVID words appear in the dataset (e.g., ``coronavirus,'' ``covid-19,'' and ``virus''), alongside traditional finance words like ``dividend.''

% \begin{figure}%
%     \centering
%     \subfloat[\centering Title Word Distributions]{{\includegraphics[width=7cm]{s2_titles_nostopwords.png} }}%
%     \qquad
%     \subfloat[\centering Description Word Distributions]{{\includegraphics[width=7cm]{s2_descrs_nostopwords.png} }}%
%     \caption{Stopword Removed Article Word Histograms}%
%     \label{s2:fig1}%
% \end{figure}

% We additionally show a visualization of aggregated article publication over time in \ref{s2:fig2}. This follows a predictable trend, where weekdays see a significant number of articles published -- up to almost 800 across all data sources -- whereas weekends dip down around the 200 mark. 

% \begin{figure}[h!]
%     \centering
%     \includegraphics[width=9cm]{s2_articles_overtime.png}
%     \caption{Number of Articles Published Over Time}
%     \label{s2:fig2}
% \end{figure}

\subsection{Further Data Analysis}

\begin{figure}[t]
    \centering
    \includegraphics[width=0.47\textwidth]{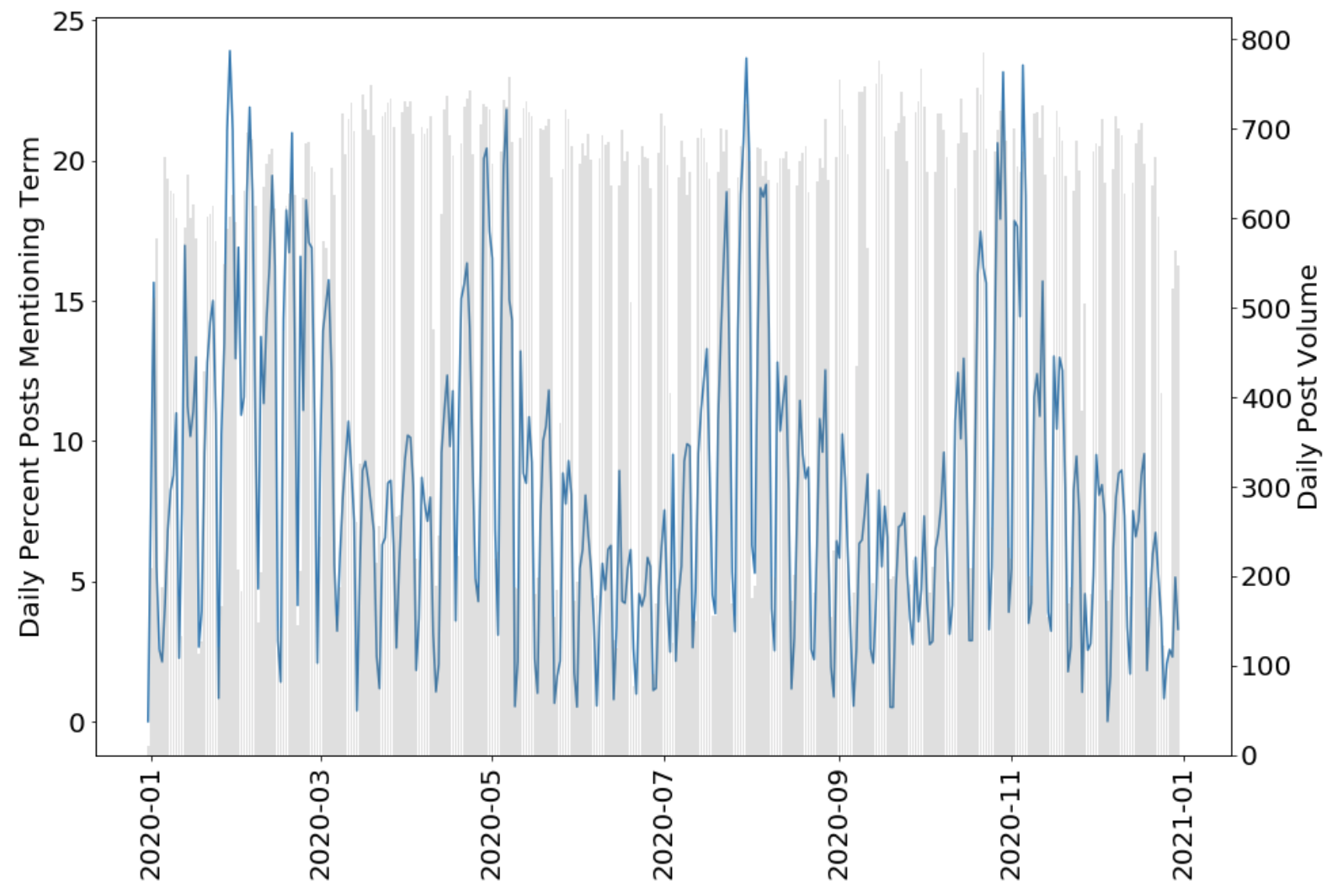}
    \caption{Percentage of posts by day corresponding to ``Earnings'' terms.}
    \label{s2:fig3}
\end{figure}

We aggregated articles by daily frequency and computed the occurrences over time for different sets of query words, producing the proportion of all posts published in a given day which contain the query words in the post title or description.  Our first plot, shown in Figure \ref{s2:fig3}, uses common ``Earnings'' query words: \{\textit{revenue}, \textit{earnings}, \textit{dividend}, \textit{miss}, \textit{beat}, \textit{estimates}\}. These words peak notably near the beginnings and ends of fiscal quarters. The blue line denotes the daily percent of posts mentioning one (or multiple) query terms, and the grey bars denote the daily post volume. 

\begin{figure}[t]
    \centering
    \includegraphics[width=0.47\textwidth]{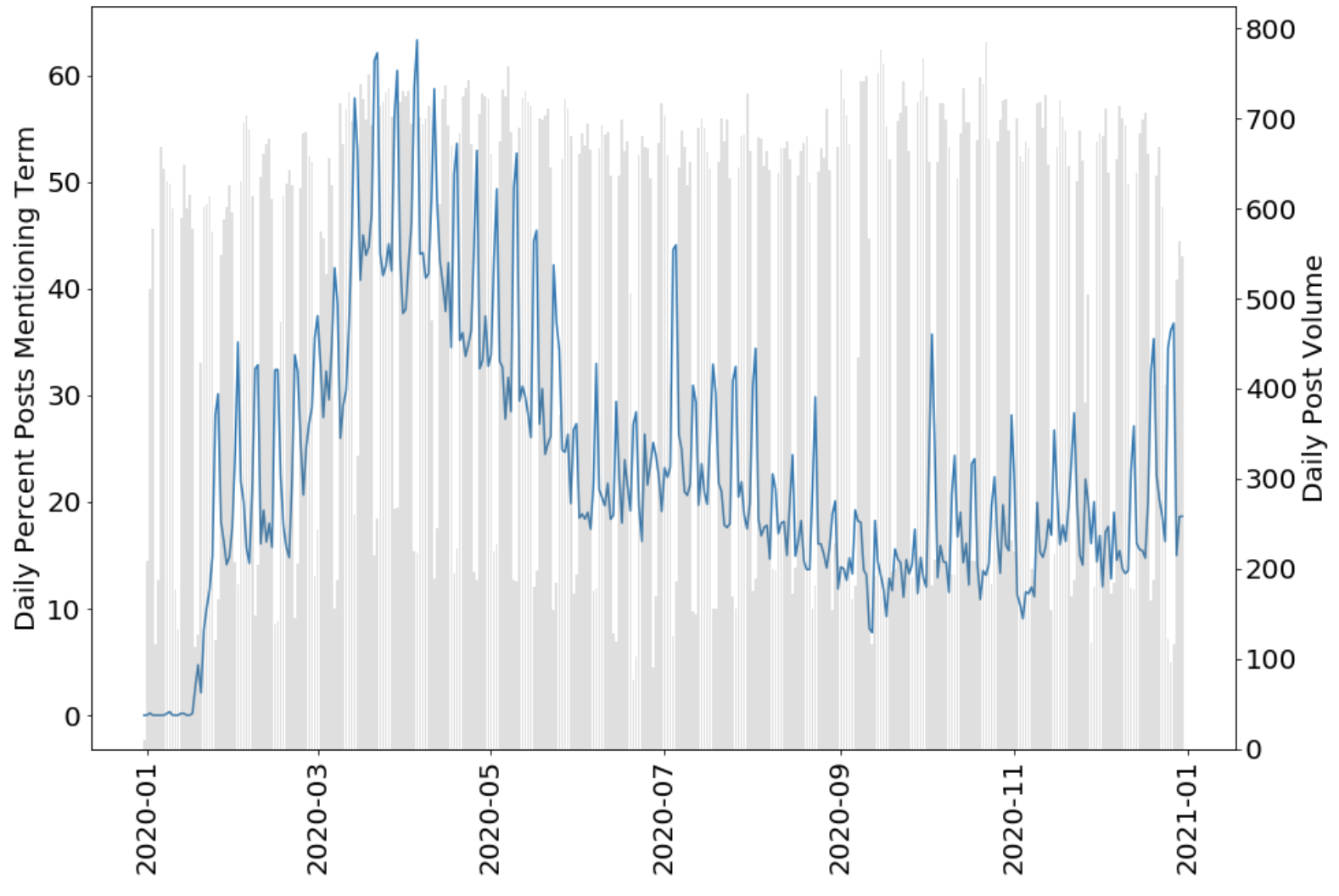}
    \caption{Percentage of posts by day corresponding to ``COVID-19-related'' terms.}
    \label{s2:fig4}
\end{figure}

We additionally consider a query of ``COVID-19-related'' words: \{\textit{covid}, \textit{covid19}, \textit{coronavirus}, \textit{virus}, \textit{corona}, \textit{covid-19}, \textit{2019-ncov}, \textit{2019ncov}, \textit{sars-cov-2}, \textit{sarscov2}\}. Interestingly, we note a spike of COVID-related keywords before the actual spike of cases in the U.S., and a steady decline between April and June despite the increasing ground truth number of cases and ongoing economic impact. One may also observe that the peak percent of posts mentioning the ``COVID-19-related'' keywords is around 60\% in late-March of 2020, and the lowest percent is in September with around 10\%. Figure \ref{s2:fig4} shows these percentages, again with the blue line denoting the daily percent of posts mentioning one (or multiple) query terms, and the grey bars denoting the daily post volume. 

When comparing to the actual COVID-19 statistics as aggregated by the U.S. Centers for Disease Control (CDC), made publicly available via their website,\footnote{\url{https://covid.cdc.gov/covid-data-tracker/}} we observe a strong initial financial press reaction to the virus, despite the low infection rate. The press decreased its focus on COVID-19 over time, despite its continuing impact on society and the economy. 

\subsection{Additional External Datasets}

While labeling the \textsc{Fin-Cov-News} data was essential for learning COVID-19 related trends, we additionally utilized samples of two external datasets created for similar tasks. For the first (\textsc{SemEval}), from SEMEVAL-2017 \cite{semeval_2017}, we sample 750 random articles and use cutoff points at -0.33 and 0.33 to threshold the continuous sentiment labels to the categorical negative, neutral, and positive. This yielded 156 negative articles, 413 neutral, and 181 positive. 

We additionally sample 750 random phrases from v1.0 of the Financial Phrase Bank \citep[\textsc{FinPhraseBank},][]{malo_good_2014}. This dataset similarly contains news phrases from financial data sources with categorical negative, neutral, and positive labels. We sampled exclusively from the ``AllAgree" subset (containing only samples exhibiting a unanimous label decision from the annotators) to ensure the highest possible quality in our training data. 

\section{Annotation}

To train our model to identify language patterns during the 2020 COVID-19 financial crisis, we first acquired labels for a sample from \textsc{Fin-Cov-News}. 

\subsection{Annotator Information}

Annotations were solicited from four annotators (one of our authors and three external experts). Our external annotators (referred to as B, C, and D) were recruited from the quantitative strategy research group of a top Investment Management/Exchange Traded Fund provider, where annotator A (author) worked for a Summer. All three external annotators are full CFA charter\footnote{\url{https://www.cfainstitute.org/en/programs/cfa}} holders and have worked in the investment space for at least three years. The experts completed the labeling task on a volunteer basis, and were given approximately two weeks to complete their labels.
 
\subsection{Labeling Scheme}
 
We selected a basic categorical (negative/neutral/positive) labeling scheme, at the article level. Annotators were provided with XLSX files with each row containing an article title, a truncated article description ($\sim$25 words), and a label column to be completed. Labeling guidelines are shown in Table \ref{s3:tab_guide}.

\begin{table}[t]
\centering
\small
\begin{tabular}{@{}lp{2.5cm}p{3cm}@{}}
\toprule
Label        & Description & Examples  \\ \midrule
Negative & Article reports news expected to negatively impact the price of a specific company or industry, or market health as a whole, either now or in the future.  & COVID-19 cases spread in a country (and it is unclear that specific company or industry stands to gain), a company announces layoffs or bankruptcy, or a news blurb indicates that the price of a share is down.   \\ \midrule
Positive & Article reports news expected to positively impact the price of a specific company or industry, or market health as a whole, either now or in the future.  & A company is in talks for a significant new business partnership, COVID-19 vaccine progress is positive, or a news blurb indicates that the price of a share is up.   \\ \midrule
Neutral & Article cannot be placed with any notable degree of certainty in positive or negative categories.  & Article mentions a company or event which has no influence on U.S.~markets, article mentions multiple companies with multiple sentiments, or annotator is uncertain if the company will benefit from or be hurt by the news.  \\ \bottomrule
\end{tabular}
\caption{Labeling guidelines for \textsc{Fin-Cov-News}.}
\label{s3:tab_guide}
\end{table}

%\begin{itemize}
%    \item \textbf{Label: Negative} -- If the text is indicating bad news which we expect will have a negative impact on the price of a specific company/industry mentioned, or market health as a whole, either now or in the future, a label of ``negative" is appropriate. 
%    \begin{itemize}
%        \item \textit{Examples}: COVID-19 cases spread in a country (and it is unclear that a specific company/industry stands to gain). A company announces layoffs or bankruptcy. A news blurb that the price of a share is down. 
%    \end{itemize}
%    \item \textbf{Label: Positive} -- If the text is indicating good news which we expect will have a positive impact on the price of a specific company/industry mentioned, or market health as a whole, either now or in the future, a label of ``positive" is appropriate. 
%    \begin{itemize}
%        \item \textit{Examples}: A company is in talks for a new business partnership which is significant in magnitude. COVID-19 vaccine progress is positive. A news blurb that the price of a specific share is up. 
%    \end{itemize}
%    \item \textbf{Label: Neutral} -- The ``neutral" category is for any article which cannot be placed with any notable degree of certainty in positive or negative categories. 
%    \begin{itemize}
%        \item \textit{Examples}: An article mentions a company or event which has no influence on U.S. markets. An article mentions multiple companies with multiple sentiments. Annotator is uncertain if the company will benefit or be hurt by the news listed. 
%    \end{itemize}
%\end{itemize}

\subsection{Inter-Annotator Agreement}

\begin{figure}[t]
    \centering
    \includegraphics[width=0.47\textwidth]{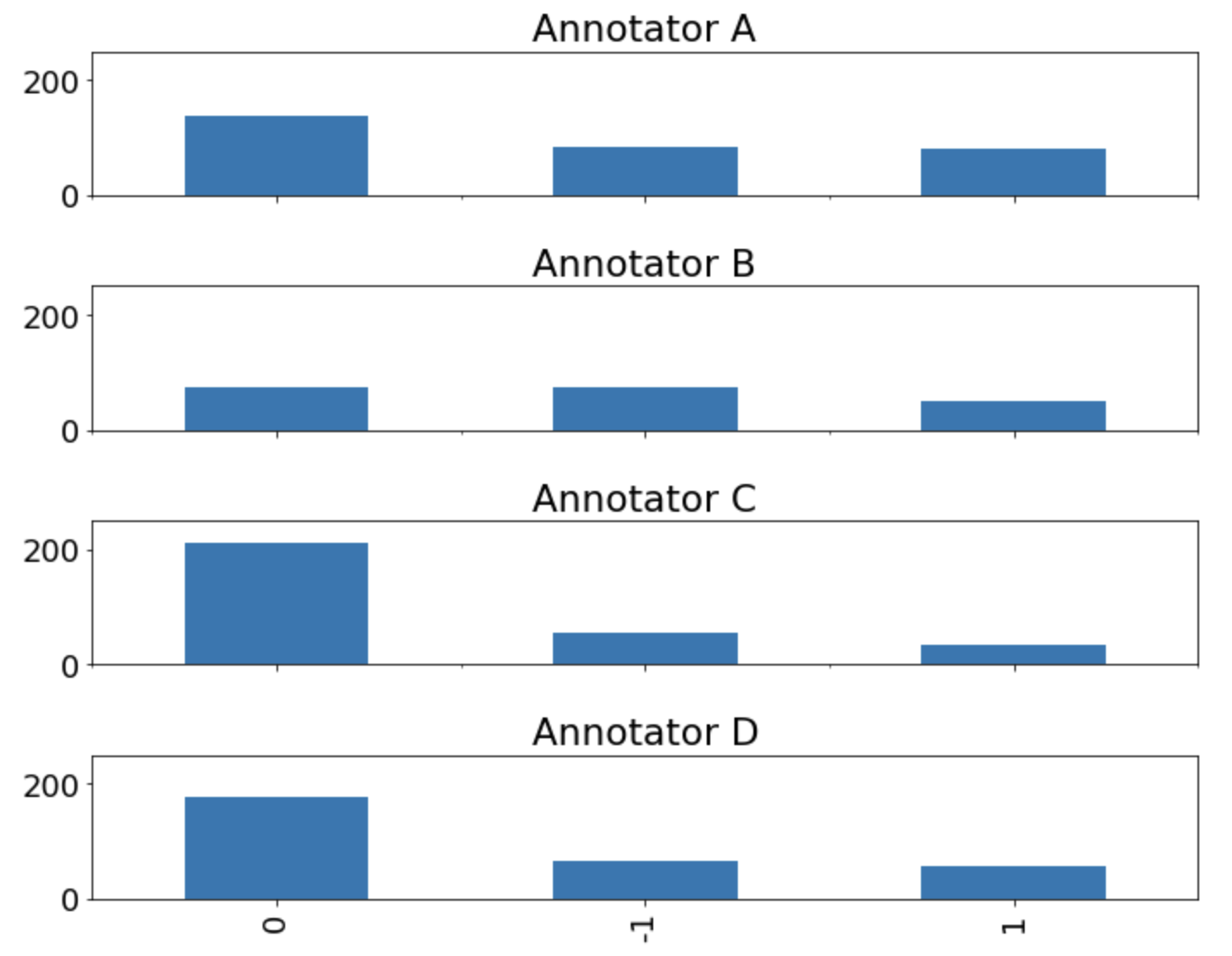}
    \caption{Annotator label distributions.}
    \label{s3:fig1}
\end{figure}

We allowed annotators to volunteer for a round number of articles they felt comfortable labeling, resulting in the following amount of data: (A, 300), (B, 200), (C, 300), (D, 300). Between each set of articles, 50 were shared among all annotators to compute inter-annotator agreement.  The distributions of all annotations, separated by annotator, are shown in Figure \ref{s3:fig1}, with -1, 0, and 1 corresponding to negative, neutral, and positive labels. Annotators C and D both leaned heavily towards assigning neutral labels and being selective with good news or bad news, whereas Annotator B had a fairly uniform distribution of labels across their data subset. 

Pairwise Cohen's Kappa \cite{cohen_coefficient_1960} scores are shown in Table \ref{s3:tab1}. Annotator B was an outlier, exhibiting poor agreement with the other two experts, C and D, likely due to the annotator's increased willingness to assign positive/negative labels. Experts C and D agreed well with each other at $\kappa=0.685$. Following convention \cite{landis_measurement_1977}, annotator pairs (A,D) and (C,D) exhibit ``substantial'' agreement, and annotator pairs (A,B) and (A,C) exhibit high ``moderate'' agreement. However, Annotator B exhibits only ``fair'' agreement with Annotator C and barely reaches the bottom threshold of ``moderate'' with Annotator D. 

Based on this analysis, we removed Annotator B from the dataset, leaving 800 labeled articles in \textsc{Fin-Cov-News}. We highlight the inherent uncertainty during the labeling of these financial documents --- upon manual inspection, it is not uncommon for ``close-misses,'' where some annotators thought an article was positive and others thought it was neutral (or some thought it was negative, others neutral). Even when considering exclusively domain experts, it is difficult to obtain complete agreement. 

\begin{table}[t]
\centering
\small
\begin{tabularx}{8cm}{*{5}{X}}
  & \textbf{A} & \textbf{B}     & \textbf{C}     & \textbf{D}     \\ \toprule
\textbf{A} & \cellcolor[HTML]{333333} & 0.574 & 0.574 & 0.709 \\ 
\textbf{B} & \cellcolor[HTML]{9B9B9B}  & \cellcolor[HTML]{333333}      & 0.305 & 0.430 \\ 
\textbf{C} & \cellcolor[HTML]{9B9B9B}  &   \cellcolor[HTML]{9B9B9B}    &  \cellcolor[HTML]{333333}     & 0.685 \\ 
\textbf{D} & \cellcolor[HTML]{9B9B9B}  &  \cellcolor[HTML]{9B9B9B}     &   \cellcolor[HTML]{9B9B9B}   & \cellcolor[HTML]{333333}  \\ \bottomrule
\end{tabularx}
\caption{Pairwise inter-annotator agreement among annotators A, B, C, and D.}
\label{s3:tab1}
\end{table}

\section{Methods and Evaluation}

We experiment with various models to predict financial sentiment, varying input data and features as well as model structure.  To evaluate our models, we report both macro and weighted F1 scores (harmonic mean of precision and recall) as computed by the Scikit-Learn Python library \cite{scikit-learn}.  We include both because although weighted F1 may more effectively reflect performance in real-world settings, it may also skew positive since we observe a slight class imbalance in our training data -- roughly 60\% of labels are neutral, with the remaining 40\% split between positive and negative.

We assessed performance relative to two baseline models, \textsc{AllNeutral} and \textsc{VADERMax}, to establish dataset learnability above chance or naive heuristics.  We report performance metrics for these baselines in Table \ref{s3:tab2}.  \textsc{AllNeutral} is a majority-class baseline that predicts ``neutral'' for each instance, and \textsc{VADERMax} is a rule-based heuristic that predicts labels based on the maximum VADER score for any word in the instance.  \textsc{AllNeutral} achieves a poor macro F1 score, but a weighted F1 score around 0.5 due to the label imbalance. \textsc{VADERMax} performs comparably, giving evidence that lexicon-based heuristics are insufficient in a financial crisis/recovery context. 

\begin{table}[t]
\centering
\small
\begin{tabular}{@{}lcc@{}}
\toprule
Method              & Macro F1 & Weighted F1 \\ \midrule
\textsc{AllNeutral} & 0.264    & 0.519    \\
\textsc{VADERMax}  & 0.300    & 0.536    \\ \bottomrule
\end{tabular}
\caption{Performance of baseline approaches.}
\label{s3:tab2}
\end{table}

\subsection{Model Design}
To ensure competitive performance, we experiment with different word embeddings, text preprocessing steps, and model inputs and architectures. All experiments were performed in Python 3, using Keras \cite{chollet2015keras}.

\begin{figure}[t]
    \centering
    \includegraphics[width=0.47\textwidth]{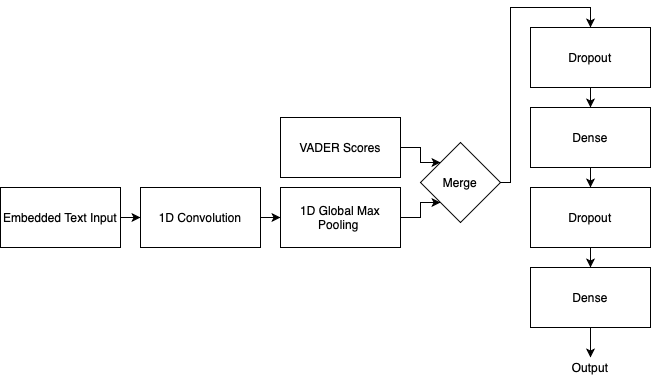}
    \caption{CNN model architecture.}
    \label{s4:fig1}
\end{figure}

For the model architecture, we adapt a similar Convolutional Neural Network (CNN) design to that of \citet{mansar_fortia-fbk_2017}, shown in Figure \ref{s4:fig1}. We also experimented with a BiGRU-based model, as they have demonstrated competitive performance relative even to Bidirectional LSTMs in comparable tasks \cite{sinha_impact_2020}. However, despite the $\sim$30x increase in training time, upon evaluation the BiGRU performed similarly to the CNN. For this reason, we selected to use the CNN for our model input experiments.\footnote{Future work may be able to achieve higher performance using more complex Transformer-based models, such as those by \citet{devlin_bert2019} and \citet{araci_finbert2019}.  Since our focus here was on establishing proof of concept in this challenging setting, substantial finetuning experiments remained out of scope but offer intriguing future possibilities.} For all model results, maximum F1 scores are reported on the testing data over 100 epochs of training.

We experimented with a variety of input features. All models are trained using a concatenation of article title text and a truncated version ($\sim$25 words) of the article description.  In this subset of experiments, we consider only the 800 labeled samples of \textsc{Fin-Cov-News} in a simple 85/15 train/test split. Our input features and preprocessing steps were as follows:
\begin{itemize}
    \item \textbf{COVID-19 Embeddings:} We created custom embeddings for the missing words \{\textit{coronavirus, covid19, 2019ncov, covid, sarscov2}\}. These embeddings are arrays of length 300, consisting of floating point numbers which would not otherwise have words assigned to them (e.g., \textit{coronavirus} = [-5.99, -5.99, ...]).
    \item \textbf{Percent Replacements:} We separated commonly seen percentages in the form of: +5\% $\rightarrow$ \{plus, 5, percent\}. 
    \item \textbf{Dash Removal:} We removed hyphens or dash characters (e.g., -, --, or ---) from the text.
    \item \textbf{Stopword Removal:} We removed stopwords using the NLTK stopwords list \cite{nltk_book}.
    \item \textbf{Punctuation Removal:} We removed all punctuation characters in the set \{!, ", \#, \$, \%, \&, ', (, ), *, +, \textbackslash{}, -, ., $/$, :, ;, $<$, =, $>$, ?, @, [, ], \textasciicircum{}, \_, `, \{, $|$, \}, $\sim$\} from the text.
\end{itemize} 

The pre-processing steps for each model in Table \ref{s4:tab1} are defined as follows: M1 - \{None\}, M2 - \{Dash Removal, COVID-19 Embeddings\}, M3 - \{Dash Removal, COVID-19 Embeddings, Stopword Removal\}, M4 - \{Dash Removal, COVID-19 Embeddings, Stopword Removal, Percent Replacements\}, M5 - \{Dash Removal, COVID-19 Embeddings, Stopword Removal, Percent Replacements, Punctuation Removal\}. We find that the optimal level of preprocessing occurs with M4.  Although the regular expressions incorporated into M5 reduce the number of missing words significantly, the model performance does not improve, and we omit the extra processing step in our final model. 

\begin{table}[t]
\centering
\small
\begin{tabular}{@{}lp{1.5cm}p{1.7cm}p{1.9cm}@{}}
\toprule
Method & Words Missing & Max Macro F1 & Max Weighted F1 \\ \midrule
M1     & 27.01\%       & 0.445        & 0.571           \\
M2     & 24.62\%       & 0.505        & 0.619           \\
M3     & 23.86\%       & 0.552        & \textbf{0.670}           \\
M4     & 22.78\%       & \textbf{0.584}        & 0.668           \\
M5     & 9.90\%        & 0.538        & 0.645           \\ \bottomrule
\end{tabular}
\caption{Performance outcomes from text preprocessing experiments, displaying maximum macro and weighted F1 scores.}
\label{s4:tab1}
\end{table}

We compare conditions leveraging pretrained GloVe embeddings, pretrained Word2Vec embeddings, and the inclusion of VADER scores in Table \ref{s4:tab2}, using a 85/15 split of \textsc{Fin-Cov-News}. The model inputs are defined as follows: N1: \{300-length GloVe 6B, VADER Scores\}, N2: \{300-length Google News Word2Vec, VADER Scores\}, N3: \{300-length Google News Word2Vec, No VADER Scores\}. We find that Word2Vec embeddings consistently result in higher performance than GloVe embeddings. This is likely because the GloVe pretraining data is more general-purpose, whereas the Word2Vec model was pretrained on news articles, likely capturing more rich relationships for our use case.  

\begin{table}[t]
\centering
\small
\begin{tabular}{@{}lp{1.5cm}p{1.7cm}p{1.9cm}@{}}
\toprule
Method & Words Missing & (Max, Avg.) Macro F1 & (Max, Avg.) Weighted F1 \\ \midrule
N1     & 17.18\%       & 0.502, 0.473 & 0.602, 0.600    \\
N2     & 22.78\%       & \textbf{0.565, 0.552} & \textbf{0.667, 0.641}    \\
N3     & 22.78\%       & 0.504, 0.483 & 0.599, 0.598    \\ \bottomrule
\end{tabular}
\caption{Performance outcomes for experiments considering word embeddings and VADER features, dislaying maximum and averaged macro and weighted F1 scores.}
\label{s4:tab2}
\end{table}

\subsection{Expansion to External Data}

For the best performing model, we expanded training to a full \textsc{SemEval} + \textsc{FinPhraseBank} + \textsc{Fin-Cov-News} dataset. The macro and weighted evaluation scores are shown in Table \ref{s4:tab3}.  Of the 345 test samples, our best performing CNN model predicted 256 articles correctly ($\sim$74.2\%), 77 off-by-one ($\sim$22.3\%), and 12 clear mispredictions ($\sim$3.5\%). 

\begin{table}[t]
\centering
\small
\begin{tabular}{@{}llll@{}}
\toprule
Method   & Precision & Recall & F1    \\ \midrule
Macro    & 0.666     & 0.679  & 0.670 \\
Weighted & 0.753     & 0.742  & 0.746 \\ \bottomrule
\end{tabular}
\caption{Performance of the best-performing model on an expanded \textsc{SemEval} + \textsc{FinPhraseBank} + \textsc{Fin-Cov-News} dataset.}
\label{s4:tab3}
\end{table}

While still an error, off-by-one errors signify that the model predicted only one label away from the true value. We further decompose these incorrect predictions to gain a better understanding of what the model was missing. We deem errors when the model erred on the side of caution (e.g., predicting neutral when the label was positive or negative) to be acceptable errors, as these errors would likely cause minimal detrimental effects in downstream systems.  We report our findings below:  
\begin{itemize}
    \item Predict neutral, true negative [acceptable error] = $14/77$ off-by-ones
    \item Predict neutral, true positive [acceptable error] = $20/77$ off-by-ones
    \item Predict negative, true neutral [unacceptable error] = $18/77$ off-by-ones
    \item Predict positive, true neutral [unacceptable error] = $25/77$ off-by-ones
\end{itemize}

Shifting perspective to the twelve instances for which the model made clear mispredictions, there were ten instances for which the model predicted positive and the true label was negative, and two instances of the opposite. Upon manual inspection of these errors, we found that the articles in question were indeed fairly ambiguous. For example, when the model saw: \textit{one best years bull market erased s\&p 500 rose almost 29 percent course 2019, good second best year bull market. took little four weeks fall apart.}, the algorithm thought this was good news -- not able to catch on the subtlety of ``little four weeks fall apart.''

%We would be remiss to not give a brief mention to more cutting-edge techniques like the many variations of BERT \cite{devlin_bert2019}, \cite{araci_finbert2019}. For the purposes of this work, since we achieved performance which rivaled our expectations based on annotator agreement and SEMEVAL benchmarks, using more advanced methods remained somewhat out of scope. 

\section{Market Correlations and Findings}

We extend the predictions of our best performing model to the entire year (January 1 -- December 31, 2020) of article titles + truncated short descriptions to uncover the relationships between financial articles and actual market movement. We remove weekends (when the market is closed) from our dataset to avoid times when share prices are relatively stable. We compute Pearson correlation coefficients in all cases using the Stats module of Scipy \cite{2020SciPy-NMeth}.

\subsection{S\& P500 Averages}
\begin{figure}[t]
    \centering
    \includegraphics[width=0.47\textwidth]{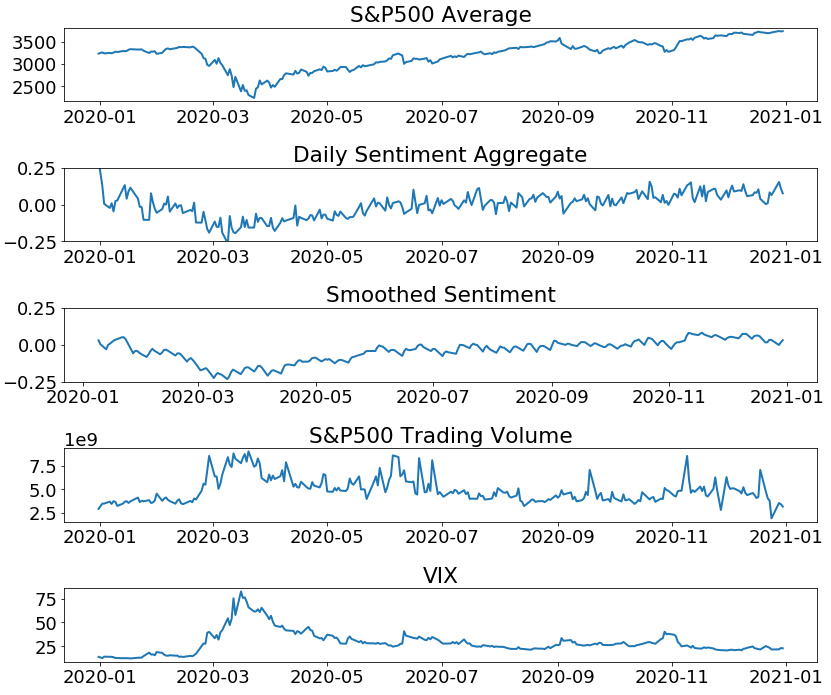}
    \caption{Stacked timeseries visualizing the S\&P 500 closing averages, averaged daily model sentiment, smoothed daily model sentiment, S\&P 500 trading volume, and Cboe Volatility Index from January 1 -- December 31, 2020.}
    \label{s5:fig1}
\end{figure}

We first compare our model predictions and the S\&P 500 average,\footnote{\url{https://finance.yahoo.com/quote/\%5EGSPC?p=\%5EGSPC}}  shown in the top two rows of Table \ref{s5:tab1}. \textit{Averaged Daily Model Sentiment} refers to a simple daily aggregate of the article sentiment. \textit{Smoothed Daily Model Sentiment} is a 10-day rolling average of the article sentiment. We find that the rolling average has a strong correlation with the S\&P 500, at $r=0.886$. This information is also presented visually in the first three plots of Figure \ref{s5:fig1}.

\begin{table}[t]
\centering
\small
\begin{tabular}{@{}m{6.5cm}r@{}}
\toprule
Item                                                  & $r$ \\ \midrule
S\&P 500 Close vs.~Averaged Daily Model Sentiment  & 0.770       \\
S\&P 500 Close vs.~Smoothed Daily Model Sentiment  & 0.886       \\
% S\&P 500 AVG Volume vs Averaged Daily Model Sentiment & -0.528      \\ 
\midrule
S\&P 500 Volume vs.~Smoothed Daily Model Sentiment & -0.563      \\
VIX Close vs.~Smoothed Daily Model Sentiment & -0.716 \\
S\&P 500 Volume vs.~VIX Close & 0.713 \\\bottomrule
\end{tabular}
\caption{Pearson correlation scores ($r$) between S\&P 500 closing averages, S\&P 500 trading volume, averaged daily model sentiment, smoothed daily model sentiment, and the Cboe Volatility Index (VIX).}
\label{s5:tab1}
\end{table}

\subsection{Trading Volume and the Cboe Volatility Index}
We additionally hypothesized that trading volume and market volatility would correlate with predicted financial news sentiment. To assess this, we considered the simple trading volume related to the S\&P500, as well as the Cboe Volatility Index (VIX) --- a well-regarded way to measure market volatility based on options contracts. These correlations are shown in rows three through five of Table \ref{s5:tab1}. Interestingly, while our smoothed sentiment model outputs exhibit non-zero correlation with both the S\&P500 trading volume and the VIX, the magnitude of this correlation is notably higher with the VIX.

\subsection{Factor Analysis}
Next, we investigated which single factor-based strategies aligned most closely with predicted financial article sentiment. Factor investing is a classical investing technique that seeks to identify certain attributes of securities that are mathematically linked with higher returns. The body of academic work on this topic is extensive (we direct interested readers to \citet{fama_common_1993,fama_five-factor_2014}).

For strategy consistency, we selected four single factor exchange traded funds (ETFs) from iShares by BlackRock,\footnote{\url{https://www.ishares.com/us/strategies/smart-beta-investing}} by far the largest ETF provider as of March 16, 2021.\footnote{\url{https://www.etf.com/sections/etf-league-tables/etf-league-table-2021-03-16}} iShares defines their four factor-based ETFs as follows:
\begin{itemize}
    \item \textbf{Value} (\textit{\$VLUE}): Stocks discounted relative to fundamentals.
    \item \textbf{Quality} (\textit{\$QUAL}): Financially healthy companies.
    \item \textbf{Momentum} (\textit{\$MTUM}): Stocks with an upward price trend.
    \item \textbf{Size} (\textit{\$SIZE}): Smaller, more nimble companies.
\end{itemize}

While these are over-simplified definitions of the factors, they provide a basic intuition regarding which companies were pooled into which of the single-factor ETFs. The statistical correlations between these tickers and our model results are shown on Table \ref{s5:tab2}. We note that Value is the clear outlier when compared to the other factors, with a significantly lower correlation ($r=0.625$). 

\begin{table}[t]
\centering
\small
\begin{tabular}{@{}m{6.5cm}r@{}}
\toprule
Item                                                  & $r$ \\ \midrule
\$VLUE Close vs.~Smoothed Daily Model Sentiment  & 0.625       \\
\$QUAL Close vs.~Smoothed Daily Model Sentiment  & 0.889       \\
\$MTUM Close vs.~Smoothed Daily Model Sentiment & 0.862      \\
\$SIZE Close vs.~Smoothed Daily Model Sentiment & 0.850      \\ \bottomrule
\end{tabular}
\caption{Pearson correlation scores ($r$) between the \$VLUE, \$QUAL, \$MTUM, and \$SIZE ETFs and smoothed daily model sentiment.}
\label{s5:tab2}
\end{table}

\subsection{FAANG+}

Finally, during the COVID-19 financial crisis and recovery, tech-focused stocks bounced back quickly and even reached all-time highs within 2020. For this reason, we specifically consider how well predicted financial news sentiment correlated with this class of stocks. For our financial data, we used the MicroSectors (BMO Harris) FAANG+ ETN (\$FNGS).\footnote{\url{https://www.microsectors.com/fang}} This ETN tracks the performance of top tech companies, weighting Facebook, Apple, Amazon, Netflix, Alphabet, Alibaba, Baidu, NVIDIA, Tesla and Twitter equally.  It displayed significantly higher returns than the overall market during our time period of interest. In \textsc{Fin-Cov-News}, 15,701 articles were tagged with one of these FAANG+ companies, accounting for around 8\% of overall articles.  

\begin{figure}[t]
    \centering
    \includegraphics[width=0.47\textwidth]{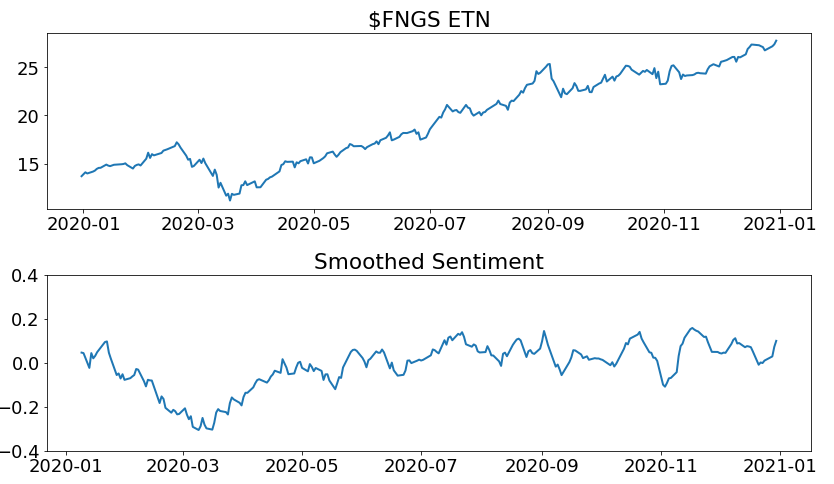}
    \caption{Stacked timeseries visualizing the \$FNGS ETN and smoothed daily model sentiment from January 1 -- December 31, 2020.}
    \label{s5:fig2}
\end{figure}

We found a correlation of $r=0.441$ (or $r=0.669$ with smoothed model sentiment) between the news for these specific companies versus the \$FNGS ETN; this was notably lower than the overall market correlations that we found earlier. Figure \ref{s5:fig2} shows that sentiment tracks market performance fairly well for the initial crash, but while the tech stocks recover quickly and add to their gains, sentiment continues to oscillate around a neutral point. This behavior highlights one of the potential shortcomings of a purely sentiment-driven trading strategy --- when market behavior is fairly stable and trending upward, the news sentiment does not always progressively grow to match this trend. 

\section{Conclusion}
In this work, we comprehensively investigated the relationship between financial news sentiment and stock market performance during the 2020 pandemic-induced financial crisis. While comparable works on this subject \cite{zammarchi_impact_2021} focus primarily on the use of social media data in conjunction with lexicons or standard machine learning models, we leveraged a task-specific CNN model to predict financial article sentiment over time, achieving a maximum F1 of 0.746. This enabled us to subsequently perform a correlation study that considered not only the stock market as a whole, but also different factor-based strategies and industry subsets.  As part of this work we collected expert annotations of financial sentiment for 800 news articles (a subset of \textsc{Fin-Cov-News}), with strong inter-annotator agreement (averaged pairwise $\kappa=0.656$).  We make these annotations available for other interested parties by request. 

With a solid deep learning-based sentiment model and an abundance of stock data, there is a high ceiling for where future work could lead. One obvious avenue would be continued improvement of the financial sentiment prediction model, although significant improvements may require more labeled data. Using the current model predictions, more fine-grained analysis could also be performed on other subsets of companies or industries of interest. While it is clear that news sentiment could not completely localize the market crash/bottom, an intriguing extension of our work would be to consider sentiment prediction on a more fine-grained time interval, and investigate whether it could be used to inform higher frequency trading strategies that could weather highly turbulent market conditions. 

% Entries for the entire Anthology, followed by custom entries
\bibliography{anthology,custom}
\bibliographystyle{acl_natbib}

\end{document}